\def\year{2019}\relax
\documentclass[letterpaper]{article} 
\usepackage{aaai19}  
\usepackage{times}  
\usepackage{helvet}  
\usepackage{courier}  
\usepackage{url}  
\usepackage{graphicx}  

\usepackage{amsmath}
\usepackage{amssymb}
\usepackage{multicol}
\usepackage{multirow}
\usepackage{booktabs}

\title{Hypergraph Neural Networks}

\author{Yifan Feng,$^1$ Haoxuan You,$^3$ Zizhao Zhang,$^3$ Rongrong Ji,$^1$$^2$ Yue Gao$^{3}$\thanks{Corresponding author. This work was finished when Yifan Feng visited Tsinghua University.}
\\
$^1$ 
Fujian Key Laboratory of Sensing and Computing for Smart City, Department of Cognitive Science\\ School of Information Science and Engineering, Xiamen University, 361005, China\\
$^2$ Peng Cheng Laboratory, China\\
$^3$BNRist, KLISS, School of Software, Tsinghua University, 100084, China.\\
{\tt\small \{evanfeng97, haoxuanyou\}@gmail.com, rrji@xmu.edu.cn, \{zz-z14,gaoyue\}@tsinghua.edu.cn}
}

\begin{document}
\maketitle
\begin{abstract}
In this paper, we present a hypergraph neural networks (HGNN) framework for data representation learning, which can encode high-order data correlation in a hypergraph structure. Confronting the challenges of learning representation for complex data in real practice, we propose to incorporate such data structure in a hypergraph, which is more flexible on data modeling, especially when dealing with complex data. In this method, a hyperedge convolution operation is designed to handle the data correlation during representation learning. In this way, traditional hypergraph learning procedure can be conducted using hyperedge convolution operations efficiently. HGNN is able to learn the hidden layer representation considering the high-order data structure, which is a general framework considering the complex data correlations. We have conducted experiments on citation network classification and visual object recognition tasks and compared HGNN with graph convolutional networks and other traditional methods. Experimental results demonstrate that the proposed HGNN method outperforms recent state-of-the-art methods. We can also reveal from the results that the proposed HGNN is superior when dealing with multi-modal data compared with existing methods.
\end{abstract}

\section{Introduction}
Graph-based convolutional neural networks \cite{kipf2016semi}, \cite{defferrard2016convolutional} have attracted much attention in recent years. Different from traditional convolutional neural networks, graph convolution is able to encode the graph structure of different input data using a neural network model and it can be used in the semi-supervised learning procedure. Graph convolutional neural networks have shown superiority on representation learning compared with traditional neural networks due to its ability of using data graph structure.

\begin{figure}
  \centering
  \includegraphics[width=3.2in]{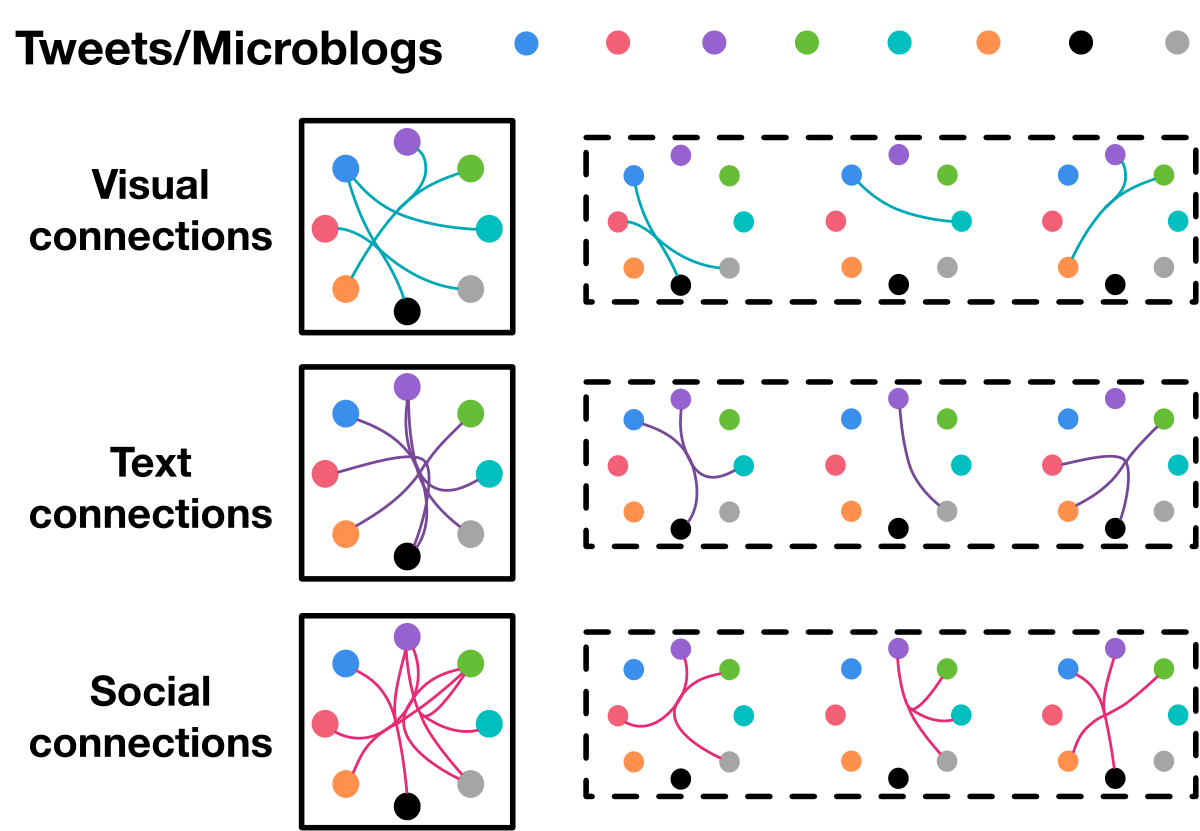}
  \caption{Examples of complex connections on social media data. Each color point represents a tweet or microblog, and there could be visual connections, text connections and social connections among them.}
\label{fig:complex_data}
\end{figure}

In traditional graph convolutional neural network methods, the pairwise connections among data are employed. It is noted that the data structure in real practice could be beyond pairwise connections and even far more complicated. Confronting the scenarios with multi-modal data, the situation for data correlation modelling could be more complex. Figure \ref{fig:complex_data} provides examples of complex connections on social media data. On one hand, the data correlation can be more complex than pairwise relationship, which is difficult to be modeled by a graph structure. On the other hand, the data representation tends to be multi-modal, such as the visual connections, text connections and social connections in this example. Under such circumstances, traditional graph structure has the limitation to formulate the data correlation, which limits the application of graph convolutional neural networks. Under such circumstance, it is important and urgent to further investigate better and more general data structure model to learn representation.

\begin{figure*}[!htbp]
  \centering
  \includegraphics[width=5in]{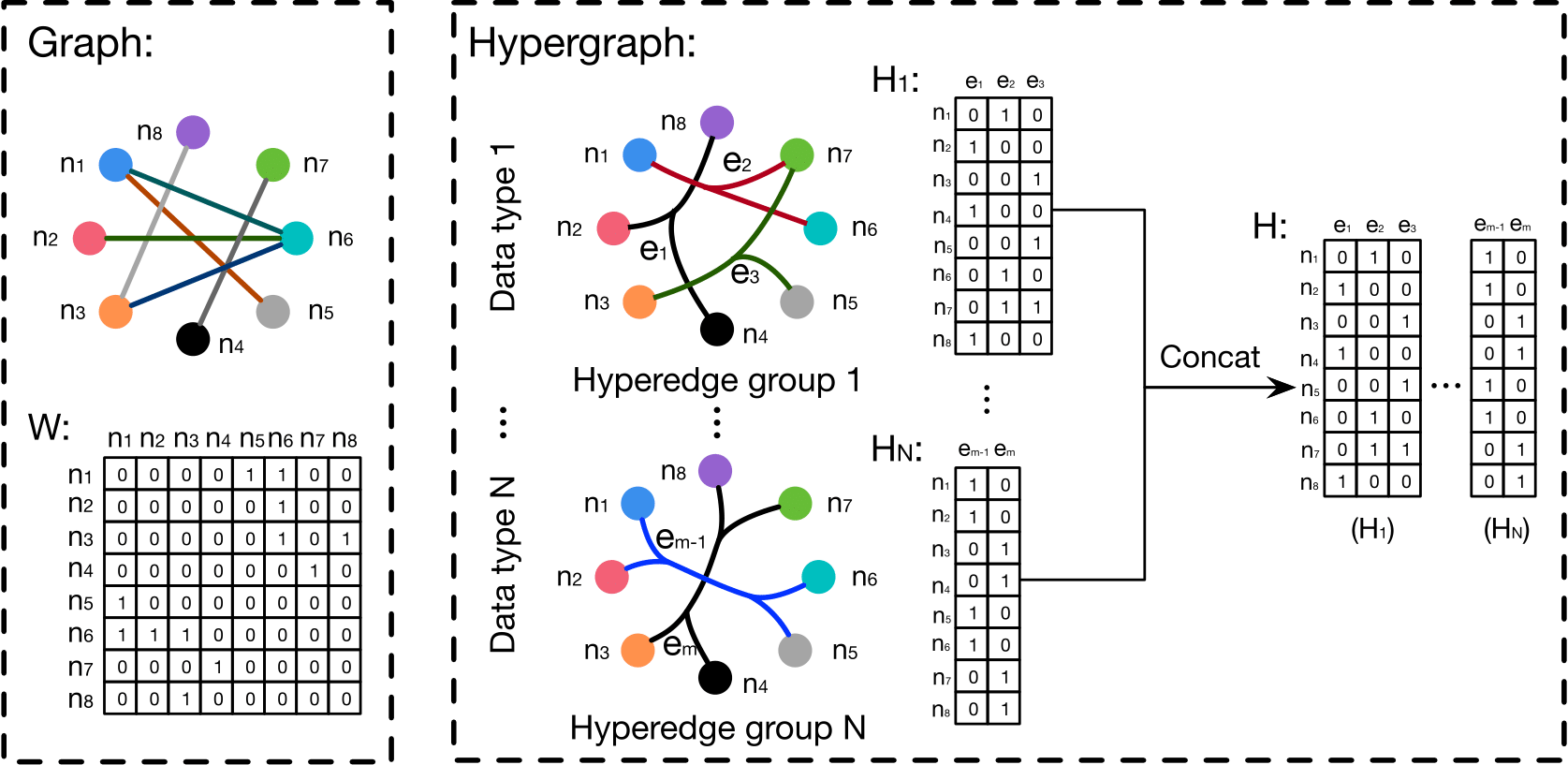}
  \caption{The comparison between graph and hypergraph.}
\label{fig:comparison_G_H}
\end{figure*}

To tackle this challenging issue, in this paper, we propose a hypergraph neural networks (HGNN) framework, which uses the hypergraph structure for data modeling. Compared with simple graph, on which the degree for all edges is mandatory 2, a hypergraph can encode high-order data correlation (beyond pairwise connections) using its degree-free hyperedges, as shown in Figure \ref{fig:comparison_G_H}. In Figure \ref{fig:comparison_G_H}, the graph is represented using the adjacency matrix, in which each edge connects just two vertices. On the contrary, a hypergraph is easy to be expanded for multi-modal and heterogeneous data representation using its flexible hyperedges. For example, a hypergraph can jointly employ multi-modal data for hypergraph generation by combining the adjacency matrix, as illustrated in Figure \ref{fig:comparison_G_H}. Therefore, hypergraph has been employed in many computer vision tasks such as classification and retrieval tasks \cite{gao20123}. However, traditional hypergraph learning methods \cite{zhou2007learning} suffer from their high computation complexity and storage cost, which limits the wide application of hypergraph learning methods.

In this paper, we propose a hypergraph neural networks framework (HGNN) for data representation learning. In this method, the complex data correlation is formulated in a hypergraph structure, and we design a hyperedge convolution operation to better exploit the high-order data correlation for representation learning. More specifically, HGNN is a general framework which can incorporate with multi-modal data and complicated data correlations. Traditional graph convolutional neural networks can be regarded as a special case of HGNN. To evaluate the performance of the proposed HGNN framework, we have conducted experiments on citation network classification and visual object recognition tasks. The experimental results on four datasets and comparisons with graph convolutional network (GCN) and other traditional methods have shown better performance of HGNN. These results indicate that the proposed HGNN method is more effective on learning data representation using high-order and complex correlations.

The main contributions of this paper are two-fold: 
\begin{enumerate}
    \item We propose a hypergraph neural networks framework, i.e., HGNN, for representation learning using hypergraph structure. HGNN is able to formulate complex and high-order data correlation through its hypergraph structure and can be also efficient using hyperedge convolution operations. It is effective on dealing with multi-modal data/features. Moreover, GCN \cite{kipf2016semi} can be regarded as a special case of HGNN, for which the edges in simple graph can be regarded as 2-order hyperedges which connect just two vertices.
    \item We have conducted extensive experiments on citation network classification and visual object classification tasks. Comparisons with state-of-the-art methods demonstrate the effectiveness of the proposed HGNN framework. Experiments also indicate the better performance of the proposed method when dealing with multi-modal data.
\end{enumerate}

\begin{figure*}[!htbp]
  \centering
  \includegraphics[width=6.6in]{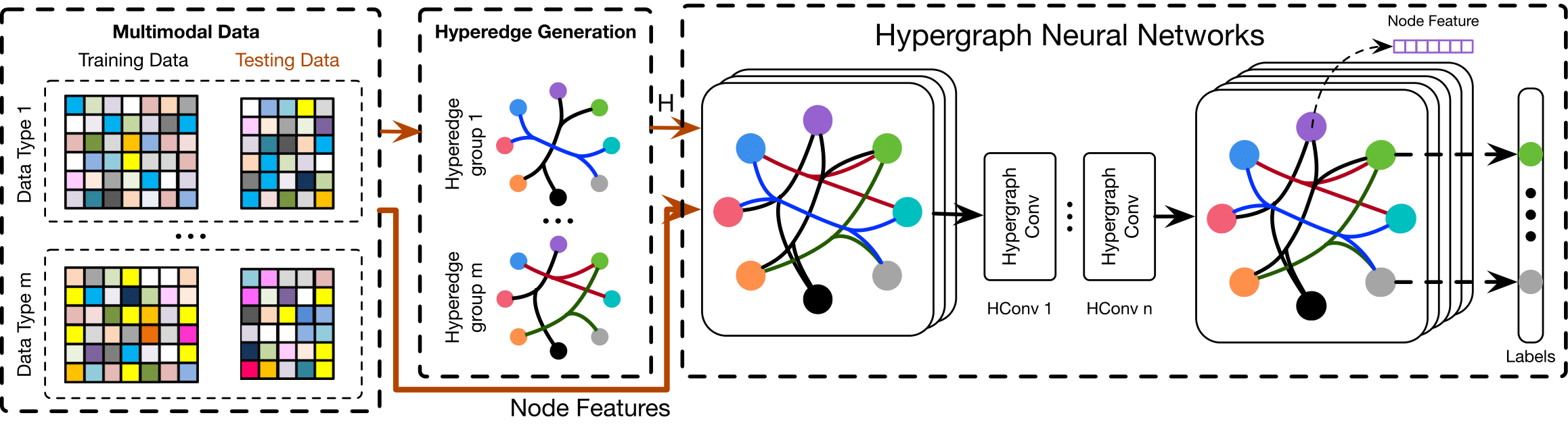}
  \caption{The proposed HGNN framework.}
\label{fig:hgnnmodel}
\end{figure*}

\section{Related Work}
\label{RW}
In this section, we briefly review existing works of hypergraph learning and neural networks on graph.

\subsection{Hypergraph learning}
In many computer vision tasks, the hypergraph structure has been employed to model high-order correlation among data. Hypergraph learning is first introduced in \cite{zhou2007learning}, as a propagation process on hypergraph structure. The transductive inference on hypergraph aims to minimize the label difference among vertices with stronger connections on hypergraph. In \cite{huang2009video}, hypergraph learning is further employed in video object segmentation. \cite{huang2010image} used the hypergraph structure to model image relationship and conducted transductive inference process for image ranking. To further improve the hypergraph structure, research attention has been attracted for leaning the weights of hyperedges, which have great influence on modeling the correlation of data. In \cite{gao2013visual}, a $\mathit{l}_2$ regularize on the weights is introduced to learn optimal hyperedge weights. In \cite{hwang2008learning}, the correlation among hyperedges is further explored by a assumption that highly correlated hyperedges should have similar weights. Regarding the multi-modal data, in \cite{gao20123}, multi-hypergraph structure is introduced to assign weights for different sub-hypergraphs, which corresponds to different modalities.

\subsection{Neural networks on graph}
Since many irregular data that do not own a grid-like structure can only be represented in the form of graph, extending neural networks to graph structure has attracted great attention from researchers. In \cite{gori2005new} and \cite{scarselli2009graph}, the neural network on graph is first introduced to apply recurrent neural networks to deal with graphs. For generalizing convolution network to graph, the methods are divided into spectral and non-spectral approaches.  

For spectral approaches, the convolution operation is formulated in spectral domain of graph. \cite{bruna2013spectral} introduces the first graph CNN, which uses the graph Laplacian eigenbasis as an analogy of the Fourier transform. In \cite{henaff2015deep}, the spectral filters can be parameterized with smooth coefficients to make them spatial-localized. In \cite{defferrard2016convolutional}, a Chebyshev expansion of the graph Laplacian is further used to approximate the spectral filters. Then, in \cite{kipf2016semi}, the chebyshev polynomials are simplified into 1-order polynomials to form an efficient layer-wise propagation model.

For spatial approaches, the convolution operation is defined in groups of spatial close nodes. In \cite{atwood2016diffusion}, the powers of a transition matrix is employed to define the neighborhood of nodes. \cite{monti2017geometric} uses the local path operators in the form of Gaussian mixture models to generalize convolution in spatial domain. In \cite{velickovic2017graph}, the attention mechanisms is introduced into the graph to build attention-based architecture to perform the node classification task on graph.

\section{Hypergraph Neural Networks}
In this section, we introduce our proposed hypergraph neural networks (HGNN). We first briefly introduce hypergraph learning, and then the spectral convolution on hypergraph is provided. Following, we analyze the relations between HGNN and existing methods. In the last part of the section, some implementation details will be given.

\subsection{Hypergraph learning statement}
We first review the hypergraph analysis theory. Different from simple graph, a hyperedge in a hypergraph connects two or more vertices. A hypergraph is defined as $\mathcal{G}=(\mathcal{V,E}, \bf W )$, which includes a vertex set $\mathcal{V}$, a hyperedge set $\mathcal{E}$. Each hyperedge is assigned with a weight by $\bf W$, a diagonal matrix of edge weights. The hypergraph $\mathcal{G}$ can be denoted by a $\vert \mathcal{V} \vert \times \vert \mathcal{E} \vert$ incidence matrix $\bf H$, with entries defined as
\begin{equation}
    h(v,e)=
    \left\{
            \begin{array}{lr}
                1, &\mbox{if}  \; v \in e\\
                0, & \mbox{if} \; v \not\in e,
            \end{array}     
    \right.
\end{equation}

For a vertex $v \in \mathcal V$, its degree is defined as $d(v)=\sum_{e\in\mathcal E}\omega(e)h(v, e)$. For an edge $e\in \mathcal{E}$, its degree is defined as $\delta(e)=\sum_{v\in\mathcal{V}}h(v,e)$. Further, ${\bf D}_e$ and ${\bf D}_v$ denote the diagonal matrices of the edge degrees and the vertex degrees, respectively.

Here let us consider the node(vertex) classification problem on hypergraph, where the node labels should be smooth on the hypergraph structure. The task can be formulated as a regularization framework as introduced by \cite{zhou2007learning}:
\begin{equation}
    \mbox{arg}\, \min_f \, \{\mathcal{R} _{emp}(f)+\Omega(f)\},
\end{equation}
where $\Omega(f)$ is a regularize on hypergraph, $\mathcal{R}_{emp}(f)$ denotes the supervised empirical loss, $f(\cdot)$ is a classification function. The regularize $\Omega(f)$ is defined as:

\begin{equation}
\begin{aligned}
    \Omega(f) =& \frac{1}{2} \sum_{e\in \mathcal{E}} \sum_{\left\{u,v\right\}\in \mathcal{V}} \frac{w(e)h(u,e)h(v,e)}{\delta(e)} \\  &\Big(\frac{f(u)}{\sqrt{d(u)}}-\frac{f(v)}{\sqrt{d(v)}}\Big) ^2,
\end{aligned}
\end{equation}

We let ${\bf \theta}={\bf D}^{-1/2}_v {\bf H}{\bf W}{\bf D}_e^{-1}{\bf H}^\top{\bf D}^{-1/2}_v$ and ${\bf \Delta}={\bf I}-{\bf \Theta}$. Then, the normalized $\Omega(f)$ can be written as
\begin{equation}
    \Omega(f) = f^\top{\bf \Delta},
\end{equation}
where $\bf \Delta$ is positive semi-definite, and usually called the hypergraph Laplacian.

\begin{figure*}[!htbp]
  \centering
  \includegraphics[width=6in]{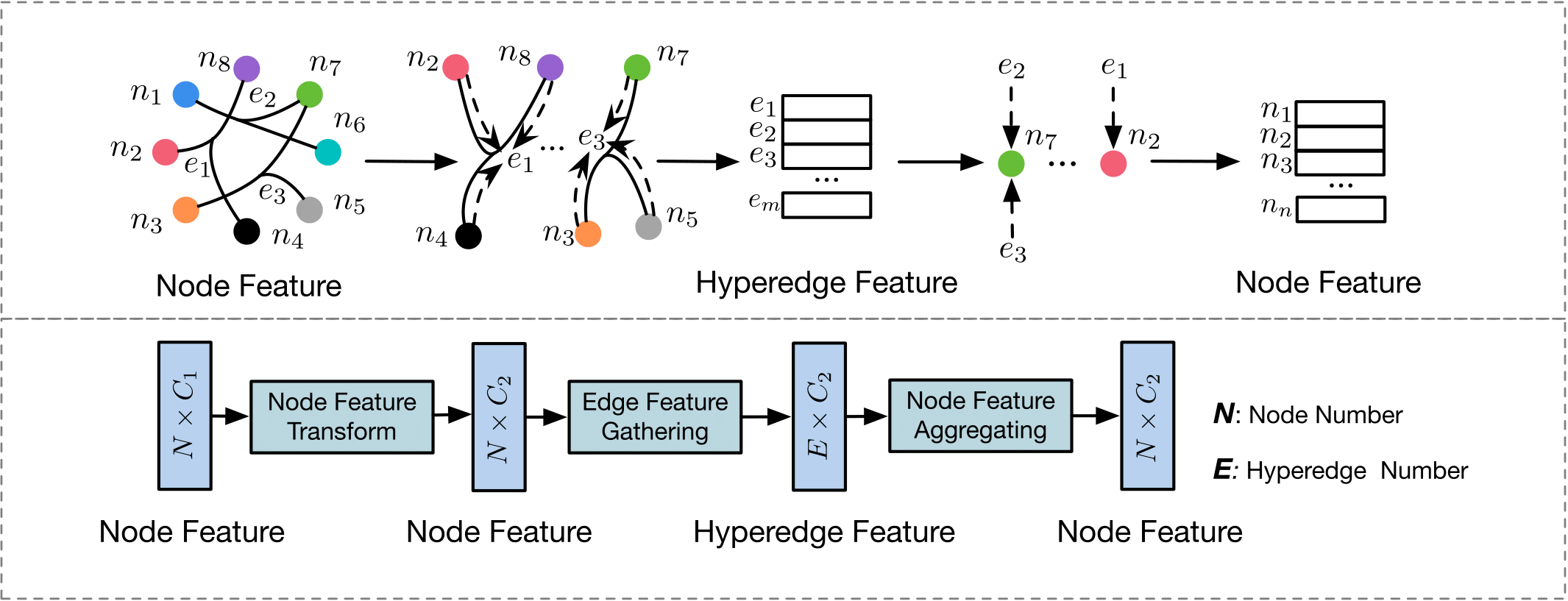}
  \caption{The illustration of the hyperedge convolution layer.}
\label{fig:hyperconv}
\end{figure*}

\subsection{Spectral convolution on hypergraph}
Given a hypergraph $\mathcal{G} = (\mathcal{V, E}, \bf \Delta)$ with $n$ vertices, since the hypergraph Laplacian ${\bf \Delta}$ is a $n \times n$ positive semi-definite matrix, the eigen decomposition ${\bf \Delta = \Phi \Lambda \Phi}^\top$ can be employed to get the orthonormal eigen vectors ${\bf \Phi} = \mbox{diag}(\phi_1, \dots, \phi_n)$ and a diagonal matrix $\bf \Lambda = \mbox{diag}(\lambda_1, \dots, \lambda_n)$ containing corresponding non-negative eigenvalues. Then, the Fourier transform for a signal $\bf x = (x_1, \dots, x_n)$ in hypergraph is defined as $\hat{\bf x}={\bf \Phi}^\top{\bf x}$, where the eigen vectors are regarded as the Fourier bases and the eigenvalues are interpreted as frequencies. The spectral convolution of signal ${\bf x}$ and filter $\bf g$ can be denoted as
\begin{equation}
    {\bf g \star x} = {\bf \Phi((\Phi^\top} {\bf g) \odot (\Phi^\top} {\bf x})) = {\bf \Phi}g({\bf \Lambda}){\bf \Phi}^\top{\bf x},
\end{equation}
where $\odot$ denotes the element-wise Hadamard product and $g(\bf \Lambda)=\mbox{diag}(g(\lambda_1), \dots, g(\lambda_n))$ is a function of the Fourier coefficients. However, the computation cost in forward and inverse Fourier transform is $\mathcal{O}(n^2)$. To solve the problem, we can follow \cite{defferrard2016convolutional} to parametrize $g(\bf \Lambda)$ with $K$ order polynomials. Furthermore, we use the truncated Chebyshev expansion as one such polynomial. Chebyshv polynomials $T_k(x)$ is recursively computed by $T_k(x)=2xT_{k-1}(x)-T_{k-2}(x)$, with $T_0(x)=1$ and $T_1(x)=x$. Thus, the $g({\bf \Lambda})$ can be parametried as
\begin{equation}
    {\bf g \star x}\approx \sum_{k=0}^{K}\theta_k T_k(\tilde{\bf \Delta}){\bf x},
    \label{conv}
\end{equation}
where $T_k(\tilde{\bf \Delta})$ is the Chebyshev polynomial of order $k$ with scaled Laplacian $\tilde{\bf \Delta} = \frac{2}{\lambda_{max}}{\bf \Delta} - \bf I$. In Equation \ref{conv}, the expansive computation of Laplacian Eigen vectors is excluded and only matrix powers, additions and multiplications are included, which brings further improvement in computation complexity. We can further let $K=1$ to limit the order of convolution operation due to that the Laplacian in hypergraph can already well represent the high-order correlation between nodes. It is also suggested in \cite{kipf2016semi} that $\lambda_{max} \approx 2$ because of the scale adaptability of neural networks. Then, the convolution operation can be further simplified to
\begin{equation}
    {\bf g \star x} \approx \theta_0 \bf x - \theta_1 {\bf D}^{-1/2}{\bf HWD}_e^{-1}{\bf H}^\top {\bf D}^{-1/2}_vx,
\end{equation}
where $\theta_0$ and $\theta_1$ is parameters of filters over all nodes. We further use a single parameter $\theta$ to avoid the overfitting problem, which is defined as
\begin{equation}
\left\{  
             \begin{array}{lr}  
			 \theta_1 = -\frac{1}{2}\theta \\
             \theta_0 = \frac{1}{2}\theta {\bf D}^{-1/2}_{v} {\bf H}{\bf D}_{e}^{-1}{\bf H}^\top{\bf D}^{-1/2}_{v},
             \end{array}  
\right.
\end{equation}

Then, the convolution operation can be simplified to the following expression
\begin{equation}
    \begin{aligned}
        {\bf g \star x} 
        &\approx \frac{1}{2}\theta {\bf D}^{-1/2}_{v} {\bf H}({\bf W}+{\bf I}){\bf D}_{e}^{-1}{\bf H}^\top{\bf D}^{-1/2}_{v}{\bf x}\\
        &\approx \theta {\bf D}^{-1/2}_{v} {\bf H}{\bf W}{\bf D}_{e}^{-1}{\bf H}^\top{\bf D}^{-1/2}_{v}{\bf x},
    \end{aligned}
\end{equation}
where $(\bf W + \bf I)$ can be regarded as the weight of the hyperedges. $\bf W$ is initialized as an identity matrix, which means equal weights for all hyperedges.

When we have a hypergraph signal $\bf X \in \mathbb{R}^{n\times C_1}$ with $n$ nodes and $C_1$ dimensional features, our hyperedge convolution can be formulated by
\begin{equation}
    \bf Y = {\bf D}^{-1/2}_{v} {\bf HWD}_e^{-1}{\bf H}^\top{\bf D}^{-1/2}_v{\bf X}\bf \Theta,
\end{equation}
where $\bf W = \mbox{diag}(w_1, \dots, w_n)$. $\Theta \in \mathbb{R}^{C_1 \times C_2}$ is the parameter to be learned during the training process. The filter $\bf \Theta$ is applied over the nodes in hypergraph to extract features. After convolution, we can obtain ${\bf Y} \in \mathbb{R}^{n\times C_2}$, which can be used for classification.

\subsection{Hypergraph neural networks analysis}
Figure \ref{fig:hgnnmodel} illustrates the details of the hypergraph neural networks. Multi-modality datasets are divided into training data and testing data, and each data contains several nodes with features. Then multiple hyperedge structure groups are constructed from the complex correlation of the multi-modality datasets. We concatenate the hyperedge groups to generate the hypergraph adjacent matrix $\bf H$. The hypergraph adjacent matrix $\bf H$ and the node feature are fed into the HGNN to get the node output labels.
As introduced in the above section, we can build a hyperedge convolutional layer $f(\bf X,W, \Theta)$ in the following formulation
\begin{equation}
{\bf X}^{(l+1)}= \sigma({\bf D}^{-1/2}_{v} {\bf H}{\bf W}{\bf D}_{e}^{-1}{\bf H}^\top{\bf D}^{-1/2}_{v}{\bf X}^{(l)}{\bf \Theta}^{(l)}),
\label{eqa:hgnn}
\end{equation}
where ${\bf X}^{(1)} \in \mathbb{R}^{N \times C}$ is the signal of hypergraph at $l$ layer, ${\bf X}^{(0)}=\bf X$ and $\sigma$ denotes the nonlinear activation function. 

The HGNN model is based on the spectral convolution on the hypergraph. Here, we further investigate HGNN in the property of exploiting high-order correlation among data. As is shown in Figure \ref{fig:hyperconv}, the HGNN layer can perform node-edge-node transform, which can better refine the features using the hypergraph structure. More specifically, at first, the initial node feature  ${\bf X}^{(1)}$ is processed by learnable filter matrix ${\bf \Theta}^{(1)}$ to extract $C_2$-dimensional feature. Then, the node feature is gathered according to the hyperedge to form the hyperedge feature $\mathbb{R}^{E\times C_2}$, which is implemented by the multiplication of $\bf H^\top \in \mathbb{R}^{E \times N}$. Finally the output node feature is obtained by aggregating their related hyperedge feature, which is achieved by multiplying matrix $\bf H$. Denote that $\bf D_v$ and $\bf D_e$ play a role of normalization in Equation \ref{eqa:hgnn}. Thus, the HGNN layer can efficiently extract the high-order correlation on hypergraph by the node-edge-node transform.

\paragraph{Relations to existing methods}
When the hyperedges only connect two vertices, the hypergraph is simplified into a simple graph and the Laplacian $\bf \Delta$ is also coincident with the Laplacian of simple graph up to a factor of $\frac{1}{2}$. Compared with the existing graph convolution methods, our HGNN can naturally model high-order relationship among data, which is effectively exploited and encoded in forming feature extraction. Compared with the traditional hypergraph method, our model is highly efficient in computation without the inverse operation of Laplacian $\bf \Delta$. It should also be noted that our HGNN has great expansibility toward multi-modal feature with the flexibility of hyperedge generation.

\subsection{Implementation}
\paragraph{Hypergraph construction}
In our visual object classification task, the features of $N$ visual object data can be represented as ${\bf X=[x_1, \dots, x_n]}^\top$. We build the hypergraph according to the distance between two features. More specifically, Euclidean distance is used to calculate $d({\bf x}_i, {\bf x}_j)$. In the construction, each vertex represents one visual object, and each hyperedge is formed by connecting one vertex and its $K$ nearest neighbors, which brings $N$ hyperedges that links $K+1$ vertices. And thus, we get the incidence matrix ${\bf H} \in \mathbb{R}^{N\times N}$ with $N \times (K+1)$ entries equaling to 1 while others equaling to 0. In the citation network classification, where the data are organized in graph structure, each hyperedge is built by linking one vertex and their neighbors according to the adjacency relation on graph. So we also get $N$ hyperedges and $\bf H \in \mathbb{R}^{N\times N}$.

\paragraph{Model for node classification}
In the problem of node classification, we build the HGNN model as in Figure \ref{fig:hgnnmodel}. The dataset is divided into training data and test data. Then hypergraph is constructed as the section above, which generates the incidence matrix $\bf H$ and corresponding $\bf D_e$. We build a two-layer HGNN model to employ the powerful capacity of HGNN layer. And the softmax function is used to generate predicted labels. During training, the cross-entropy loss for the training data is back-propagated to update the parameters $\bf \Theta$ and in testing, the labels of test data is predicted for evaluating the performance. When there are multi-modal information incorporate them by the construction of hyperedge groups and then various hyperedges are fused together to model the complex relationship on data.

\section{Experiments}
\label{exp}
In this section, we evaluate our proposed hypergraph neural  networks on two task: citation network classification and visual object recognition. We also compare the proposed method with graph convolutional networks and other state-of-the-art methods.

\begin{table}[!htbp]
    \centering
    \begin{tabular}{c|c|c}
        \toprule
         Dataset & Cora & Pumbed \\
         \midrule
         Nodes & 2708 & 19717\\
         Edges& 5429 & 44338 \\
         Feature & 1433 & 500 \\
         Training node& 140 & 60 \\
         Validation node& 500 & 500 \\
         Testing node& 1000 & 1000 \\
         Classes& 7 & 3 \\
         \bottomrule
    \end{tabular}
    \caption{Summary of the citation classification datasets.}
    \label{tab:citation_dataset}
\end{table}

\subsection{Citation network classification}
\paragraph{Datasets}
In this experiment, the task is to classify citation data. Here, two widely used citation network datasets, i.e., Cora and Pubmed \cite{sen2008collective} are employed. The experimental setup follows the settings in \cite{yang2016revisiting}. In both of those two datasets, the feature for each data is the bag-of-words representation of documents. The data connection, i.e., the graph structure, indicates the citations among those data. To generate the hypergraph structure for HGNN, each time one vertex in the graph is selected as the centroid and its connected vertices are used to generate one hyperedge including the centroid itself. Through this we can obtain the same size incidence matrix compared with the original graph. It is noted that as there are no more information for data relationship, the generated hypergraph constructure is quite similar to the graph. The Cora dataset contains 2708 data and 5\% are used as labeled data for training. The Pubmed dataset contains 19717 data, and only 0.3\% are used for training. The detailed description for the two datasets listed in Table \ref{tab:citation_dataset}.

\paragraph{Experimental settings}
In this experiment, a two-layer HGNN is applied. The feature dimension of the hidden layer is set as 16 and the dropout \cite{srivastava2014dropout} is employed to avoid overfitting with drop rate $p=0.5$. We choose the ReLU as the nonlinear activation function. During the training process, we use Adam optimizer \cite{kingma2014adam} to minimize our cross-entropy loss function with a learning rate of 0.001. We have also compared the proposed HGNN with recent methods in these experiments.

\paragraph{Results and discussion}
The results of the experimental results and comparisons on the citation network dataset are shown in Table \ref{tab:citation_result}. For our HGNN model, we report the average classification accuracy of 100 runs on Core and Pumbed, which is 81.6\% and 80.1\%. As shown in the results, the proposed HGNN model can achieve the best or comparable performance compared with the state-of-the-art methods. Compared with GCN, the proposed HGNN method can achieve a slight improvement on the Cora dataset and 1.1\% improvement on the Pubmed dataset. We note that the generated hypergraph structure is quite similar to the graph structure as there is neither extra nor more complex information in these data. Therefore, the gain obtained by HGNN is not very significant.

\begin{table}[!htbp]
  \centering
  \begin{tabular}{p{1.7in}cc}
    \toprule
      Method & Cora  & Pubmed  \\
    \midrule
    DeepWalk \cite{perozzi2014deepwalk} & 67.2\%  & 65.3\%  \\
    ICA \cite{lu2003link}  & 75.1\%  & 73.9\%  \\
    Planetoid \cite{yang2016revisiting}  & 75.7\%  & 77.2\%  \\
    Chebyshev \cite{defferrard2016convolutional} & 81.2\%  & 74.4\%  \\
    GCN \cite{kipf2016semi} & 81.5\%  & 79.0\%  \\
    \midrule
    HGNN & \textbf{81.6}\%  & \textbf{80.1}\%  \\
    \bottomrule
  \end{tabular}
  \caption{Classification results on the Cora and Pubmed datasets.}
  \label{tab:citation_result}
\end{table}

\subsection{Visual object classification}
\paragraph{Datasets and experimental settings} In this experiment, the task is to classify visual objects. Two public benchmarks are employed here, including the Princeton ModelNet40 dataset \cite{wu20153d} and the National Taiwan University (NTU) 3D model dataset \cite{chen2003visual}, as shown in Table \ref{tab:visual_dataset}. The ModelNet40 dataset consists of 12,311 objects from 40 popular categories, and the same training/testing split is applied as introduced in \cite{wu20153d}, where 9,843 objects are used for training and 2,468 objects are used for testing. The NTU dataset is composed of 2,012 3D shapes from 67 categories, including car, chair, chess, chip, clock, cup, door, frame, pen, plant leaf and so on. In the NTU dataset, 80\% data are used for training and the other 20\% data are used for testing. In this experiment, each 3D object is represented by the extracted features. Here, two recent state-of-the-art shape representation methods are employed, including Multi-view Convolutional Neural Network (MVCNN) \cite{su2015multi} and Group-View Convolutional Neural Network (GVCNN) \cite{feng2018gvcnn}. These two methods are selected due to that they have shown satisfactory performance on 3D object representation. We follow the experimental settings of MVCNN and GVCNN to generate multiple views of each 3D object. Here, 12 virtual cameras are employed to capture views with a interval angle of 30 degree, and then both the MVCNN and the GVCNN features are extracted accordingly.

To compare with GCN method, it is noted that there is no available graph structure in the ModelNet40 dataset and the NTU dataset. Therefore, we construct a probability graph based on the distance of nodes. Given the features of data, the affinity matrix $A$ is generated to represent the relationship among different vertices, and $A_{ij}$ can be calculated by:
\begin{equation}
    A_{ij}=\exp(-\frac{2D{ij}^2}{\Delta})  
    \label{eq:graph_adj}
\end{equation}
where $D_{ij}$ indicates the Euclidean distance between node $i$ and node $j$. $\Delta $ is the average pairwise
distance between nodes. For the GCN experiment with two features constructed simple graphs, we simply average the two modality adjacency matrices to get the fused graph structure for comparison.

\begin{table}[!htbp]
    \centering
    \begin{tabular}{c|c|c}
        \toprule
         Dataset & ModelNet40 & NTU \\
         \midrule
         Objects & 12311 & 2012\\
         MVCNN Feature & 4096 & 4096 \\
         GVCNN Feature & 2048 & 2048 \\
         Training node& 9843 & 1639 \\ 
         Testing node& 2468 & 373 \\
         Classes& 40 & 67 \\
         \bottomrule
    \end{tabular}
    \caption{The detailed information of the ModelNet40 and the NTU datasets.}
    \label{tab:visual_dataset}
\end{table}

\begin{table*}[!htbp]
   \centering
  \begin{tabular}{c|c|c|c|c|c|c}
    \toprule
    \multirow{3}{*}{Feature}  & \multicolumn{6}{c}{Features for Structure} \\
    \cline{2-7}
    & \multicolumn{2}{c|}{GVCNN} & \multicolumn{2}{c|}{MVCNN} & \multicolumn{2}{c}{GVCNN+MVCNN} \\
    \cline{2-7}
    & GCN & HGNN & GCN & HGNN & GCN & HGNN \\
    \midrule
    GVCNN \cite{feng2018gvcnn} & $91.8\%$ & $\textbf{92.6}\%$ & $91.5\%$ & $\textbf{91.8}\%$ & $92.8\%$ & $\textbf{96.6}\%$ \\
    MVCNN \cite{su2015multi} & $92.5\%$ & $\textbf{92.9}\%$ & $86.7\%$ & $\textbf{91.0}\%$ & $92.3\%$ & $\textbf{96.6}\%$  \\
    GVCNN+MVCNN & - & - & - & - & $94.4\%$ & \textbf{96.7$\%$} \\
    \bottomrule
  \end{tabular}
  \caption{Comparison between GCN and HGNN on the ModelNet40 dataset.}
  \label{tab:modelnet40_cmp}
\end{table*}

\begin{table*}[!htbp]
   \centering
  \begin{tabular}{c|c|c|c|c|c|c}
    \toprule
    \multirow{3}{*}{Feature}  & \multicolumn{6}{c}{Features for Structure} \\
    \cline{2-7}
    & \multicolumn{2}{c|}{GVCNN} & \multicolumn{2}{c|}{MVCNN} & \multicolumn{2}{c}{GVCNN+MVCNN} \\
    \cline{2-7}
    & GCN & HGNN & GCN & HGNN & GCN & HGNN \\
    \midrule
    GVCNN (\cite{feng2018gvcnn}) & $78.8\%$ & $\textbf{82.5}\%$ & $78.8\%$ & $\textbf{79.1}\%$ & $75.9\%$ & $\textbf{84.2}\%$ \\
    MVCNN (\cite{su2015multi}) & $74.0\%$ & $\textbf{77.2}\%$ & $71.3\%$ & $\textbf{75.6}\%$ & $73.2\%$ & $\textbf{83.6}\%$  \\
    GVCNN+MVCNN & $-$ & $-$ & $-$ & $-$ & $76.1\%$ & \textbf{84.2$\%$} \\
    \bottomrule
  \end{tabular}
  \caption{Comparison between GCN and HGNN on the NTU dataset.}
  \label{tab:ntu_cmp}
\end{table*}

\begin{table}[!htbp]
    \centering
    \begin{tabular}{c|c}
        \toprule
        \multirow{2}{*}{Method} & Classification \\
        & Accuracy \\
        \midrule
         PointNet \cite{qi2017pointnet} & 89.2\% \\
         PointNet++ \cite{qi2017pointnet++}& 90.7\%\\
         PointCNN \cite{li2018pointcnn} & 91.8\%\\
         SO-Net \cite{li2018so} & 93.4\% \\
         HGNN & \textbf{96.7\%} \\
         \bottomrule
    \end{tabular}
    \caption{Experimental comparison among recent classification methods on ModelNet40 dataset.}
    \label{tab:modelnet40_cls}
\end{table}

\begin{figure}[!htbp]
  \centering
  \includegraphics[width=3.2in]{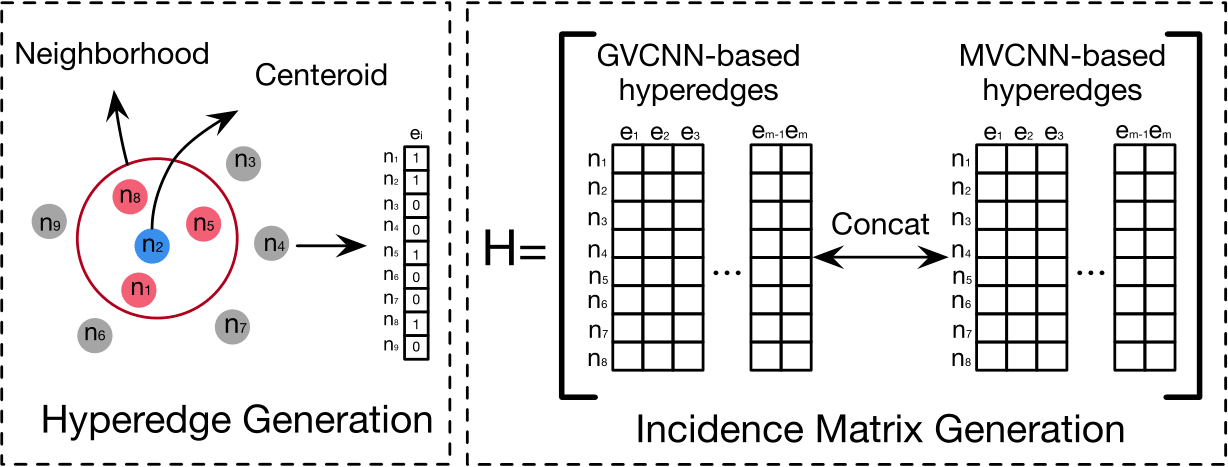}
  \caption{An example of hyperedge generation in the visual object classification task. Left: For each node we aggregate its $N$ neighbor nodes by Euclidean distance to generate a hyperedge. Right: To generate the multi-modality hypergraph adjacent matrix we concatenate adjacent matrix of two modality.}
\label{fig:edge_gene}
\end{figure}

\subsubsection{Hypergraph structure construction on visual datasets}
In experiments on ModelNet40 and NTU datasets, two hypergraph construction methods are employed. The first one is based on single modality feature and the other one is based on multi-modality feature. In the first case, only one feature is used. Each time one object in the dataset is selected as the centroid, and its 10 nearest neighbors in the selected feature space are used to generate one hyperedge including the centroid itself, as shown in Figure \ref{fig:edge_gene}. Then, a hypergraph $\mathcal{G}$ with $N$ hyperedges can be constructed. In the second case, multiple features are used to generate a hypergraph $\mathcal{G}$ modeling complex multi-modality correlation. Here, for the $i^{th}$ modality data,  a hypergraph adjacent matrix $\bf{H}_i$ is constructed accordingly. After all the hypergraphs from different features have been generated, these adjacent matrices $\bf H_i$ can be concatenated to build the multi-modality hypergraph adjacent matrix $\bf H$. In this way, the hypergraphs using single modal feature and multi-modal features can be constructed.

\paragraph{Results and discussions}
Experiments and comparisons on the visual object recognition task are shown in Table \ref{tab:modelnet40_cmp} and Table \ref{tab:ntu_cmp}, respectively. For the ModelNet40 dataset, we have  compared the proposed method using two features with recent state-of-the-are methods in Table \ref{tab:modelnet40_cls}. As shown in the results, we can have the following observations:
\begin{enumerate}
    \item The proposed HGNN method outperforms the state-of-the-art object recognition methods in the ModelNet40 dataset. More specifically, compared with PointCNN and SO-Net, the proposed HGNN method can achieve gains of 4.8\% and 3.2\%, respectively. These results demonstrate the superior performance of the proposed HGNN method on visual object recognition.
    \item Compared with GCN, the proposed method achieves better performance in all experiments. As shown in Table \ref{tab:modelnet40_cmp} and Table \ref{tab:ntu_cmp}, when only one feature is used for graph/hypergraph structure generation, HGNN can obtain slightly improvement. For example, when GVCNN is used as the object feature and MVCNN is used for graph/hypergraph structure generation, HGNN achieves gains of 0.3\% and 2.0\% compared with GCN on the ModelNet40 and the NTU datasets, respectively. When more features, i.e., both GVCNN and MVCNN, are used for graph/hypergraph structure generation, HGNN achieves much better performance compared with GCN. For example, HGNN achieves gains of 8.3\%, 10.4\% and 8.1\% compared with GCN when GVCNN, MVCNN and GVCNN+MVCNN are used as the object features on the NTU dataset, respectively.
\end{enumerate}

The better performance can be dedicated to the employed hypergraph structure. The hypergraph structure is able to convey complex and high-order correlations among data, which can better represent the underneath data relationship compared with graph structure or the methods without graph structure. Moreover, when multi-modal data/features are available, HGNN has the advantage of combining such multi-modal information in the same structure by its flexible hyperedges. Compared with traditional hypergraph learning methods, which may suffer from the high computational complexity and storage cost, the proposed HGNN framework is much more efficient through the hyperedge convolution operation.

\section{Conclusion}
In this paper, we propose a framework of hypergraph neural networks (HGNN). In this method, HGNN generalizes the convolution operation to the hypergraph learning process. The convolution on spectral domain is conducted with hypergraph Laplacian and further approximated by truncated chebyshev polynomials. HGNN is a more general framework which is able to handle the complex and high-order correlations through the hypergraph structure for representation learning compared with traditional graph. We have conducted experiments on citation network classification and visual object recognition tasks to evaluate the performance of the proposed HGNN method. Experimental results and comparisons with the state-of-the-art methods demonstrate better performance of the proposed HGNN model.  HGNN is able to take complex data correlation into representation learning and thus lead to potential wide applications in many tasks, such as visual recognition, retrieval and data classification.


\section*{Acknowledgements}
This work was supported by National Key R\&D Program of China (Grant No. 2017YFC0113000, and No.2016YFB1001503), and National Natural Science Funds of China (No.U1705262, No.61772443, No.61572410, No.61671267),
National Science and Technology Major Project (No.
2016ZX01038101), MIIT IT funds (Research and application
of TCN key technologies) of China, and The National Key
Technology R and D Program (No. 2015BAG14B01-02), Post Doctoral Innovative Talent Support Program under Grant BX201600094, China Post-Doctoral Science Foundation under Grant 2017M612134, Scientific Research Project of National Language Committee of China (Grant No. YB135-49), and Nature Science Foundation of Fujian Province, China (No. 2017J01125 and No. 2018J01106).

\bibliographystyle{aaai}
\bibliography{AAAI-FengY.1506.bib}

\end{document}